\UseRawInputEncoding
\pdfoutput=1
\documentclass[journal]{IEEEtran}
\usepackage{algorithm}
\usepackage{algorithmic}
\usepackage{graphicx}
\graphicspath{{figures/}}
\usepackage{textcomp}
\RequirePackage{CJKnumb}

\usepackage{bm}
\usepackage[tight,footnotesize]{subfigure}

\usepackage{url}

\usepackage{multirow}
\usepackage{longtable}
\usepackage{stfloats}
\usepackage{color}

\ifCLASSINFOpdf
\else
\fi

\usepackage{amssymb}
\usepackage[cmex10]{amsmath}
\begin{document}

\title{SAR-to-Optical Image Translation via Thermodynamics-inspired Network}

\author{Mingjin~Zhang,~\IEEEmembership{Member,~IEEE},
        Jiamin~Xu,
        Chengyu~He,
        Wenteng~Shang,
        Yunsong~Li,     and~Xinbo~Gao,~\IEEEmembership{Senior~Member,~IEEE}
\thanks{M. Zhang, J. Xu, C. He, W. Shang, and Y. Li are with the State Key Laboratory of Integrated Services Networks, Xidian University, Xi'an 710071, Shaanxi, P. R. China.

X. Gao (Corresponding author)  is with the Chongqing Key Laboratory of Image Cognition, Chongqing University of Posts and Telecommunications, Chongqing 400065, China (e-mail: gaoxb@cqupt.edu.cn), and with the School of Electronic Engineering, Xidian University, Xi’an 710071, China (e-mail: xbgao@mail.xidian.edu.cn).}
}
\maketitle

\begin{abstract}
Synthetic aperture radar (SAR) is prevalent in the remote sensing field but is difficult to interpret in human visual perception. Recently, SAR-to-optical (S2O) image conversion methods have provided a prospective solution for interpretation.
However, since there is a huge domain difference between optical and SAR images, they suffer from low image quality and geometric distortion in the produced optical images. Motivated by the analogy between pixels during the S2O image translation and molecules in a heat field, Thermodynamics-inspired Network for SAR-to-Optical Image Translation (S2O-TDN) is proposed in this paper. Specifically, we design a Third-order Finite Difference (TFD) residual structure in light of the TFD equation of thermodynamics, which allows us to efficiently extract inter-domain invariant features and facilitate the learning of the nonlinear translation mapping. In addition, we exploit the first law of thermodynamics (FLT) to devise an FLT-guided branch that promotes the state transition of the feature values from the unstable diffusion state to the stable one, aiming to regularize the feature diffusion and preserve image structures during S2O image translation. S2O-TDN follows an explicit design principle derived from thermodynamic theory and enjoys the advantage of explainability. Experiments on the public SEN1-2 dataset show the advantages of the proposed S2O-TDN over the current methods with more delicate textures and higher quantitative results.

\end{abstract}


\IEEEpeerreviewmaketitle

\section{Introduction}

Synthetic aperture radar (SAR) achieves high-resolution microwave imaging by recording the amount of energy reflected from the sensor's emitted energy~\cite{2021SAR} as it interacts with the Earth. Since the wavelength range of the electromagnetic signals used by SAR is centimeter to meter, SAR can effectively penetrate concealed objects and clouds. In addition, SAR is not affected by illumination conditions and can operate under all-weather, all-day conditions, which is currently an important means of remote sensing observation in many applications~\cite{2021An,huang2020underwater,argenti2011simplified}. Nevertheless, they are more difficult to interpret by human vision than optical images. Therefore, many works have been proposed to achieve SAR-to-Optical (S2O) image translation~\cite{merkle2018exploring,wang2018generating} as well as to improve the readability of SAR images. However, geometric distortions in SAR images~\cite{auer2009ray} are unavoidable, resulting from the special imaging mechanism of SAR images. In addition, the consistent interference of object scattering on radar echoes makes certain pixel diffusion problems inevitable during imaging. Considering the large difference in content between SAR and optical images, learning the desired nonlinear mapping of SAR images to optical images during S2O image translation remains a challenge.

\begin{figure}[!t]
\centering
\includegraphics[width=1.0\columnwidth]{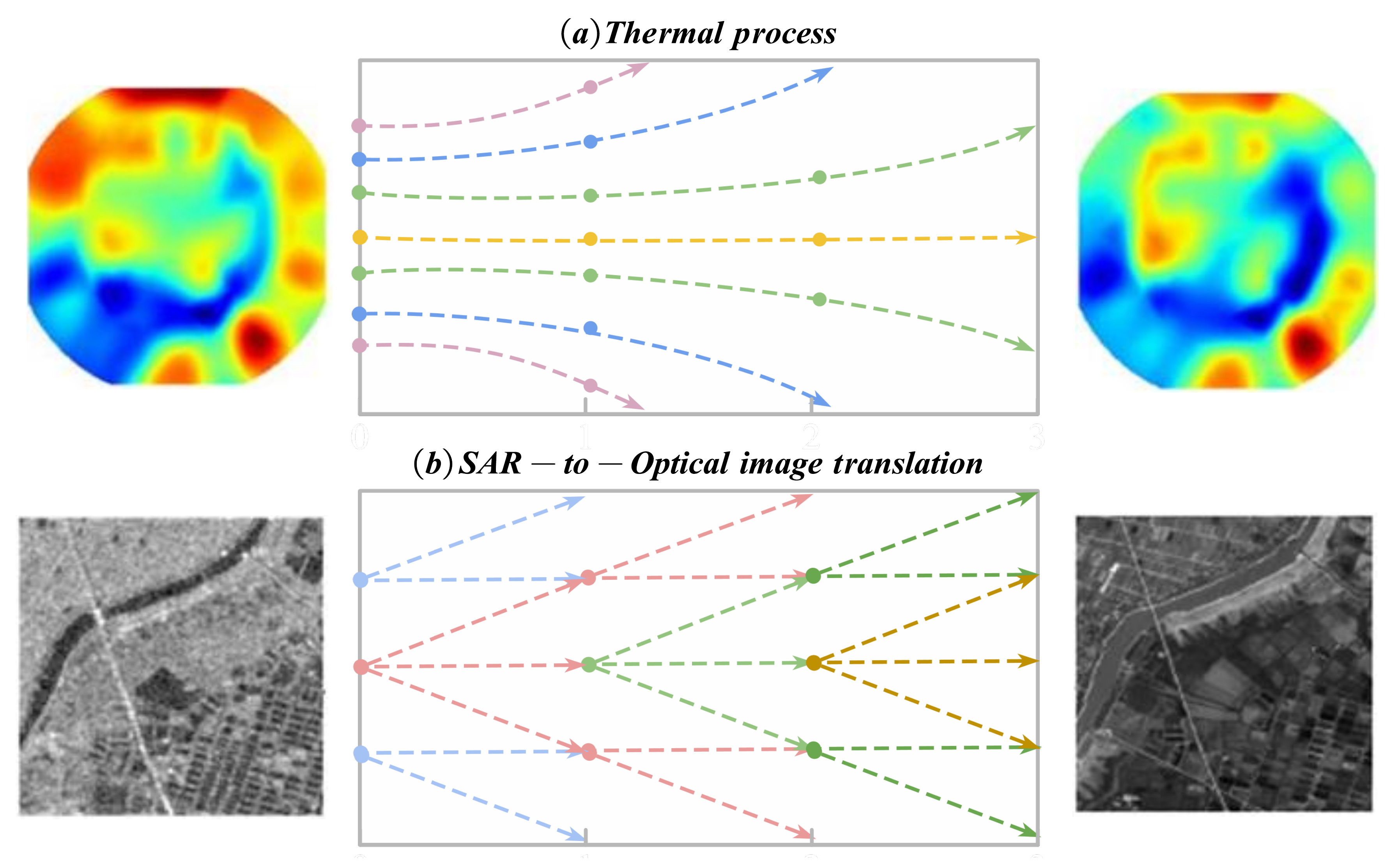}  
\caption{Analogy between (a) thermal field of thermodynamics, where the particles collide and the heat state transforms from unstable to stable, and (b) S2O image translation process, where the image is rearranged with features as the pixel moves forward from the microscopic perspective.}
\label{1}
\end{figure}

As shown in Figure~\ref{1}, particles in the closed thermal field described in thermodynamics, are in frequent collision and motion under the constraint of temperature difference~\cite{laine2016basics}. The heat is simultaneously absorbed or released as it moves from the unstable high-temperature state at the left of the figure to the more stable low-temperature state at the right of the figure. On the other hand, during the S2O image conversion, pixels with different values of a SAR image are shifted and diffused within the objective function constraint, whose values can be varied based on the optical image for reconstruction. Analogously, the above two processes share some similarities in the behavior of individual particle (pixel) movement, suggesting that some of the theories of thermodynamics can be adapted to the S2O image translation process. For example, due to the difficulty of obtaining exact numerical solutions for continuous variables, the first-order difference equations of thermodynamics are utilized to achieve numerical solutions for particles at discrete time points. It can be adopted in the same way as the expression for the residual network. Given that higher-order residual networks have a greater learning capability, we can use the third-order equation instead of the first-order finite difference equation to guide the construction of the residual structure to obtain a more efficient function. In addition, in the S2O image translation process, when extracting SAR image features as time changes, the network is in an unstable state due to the uncertainty of pixel motion, resulting in the pixels undergoing diffusion. In light of the first law of thermodynamics~\cite{romano2010variational},  the thermal field tends to move from an unstable higher temperature state to a more stable lower temperature state. Therefore, we consider following the first law of thermodynamics and designing a guided branch to regularize the feature diffusion.

\begin{figure*}[t!]
	\centering
	\includegraphics[width=1.0\linewidth]{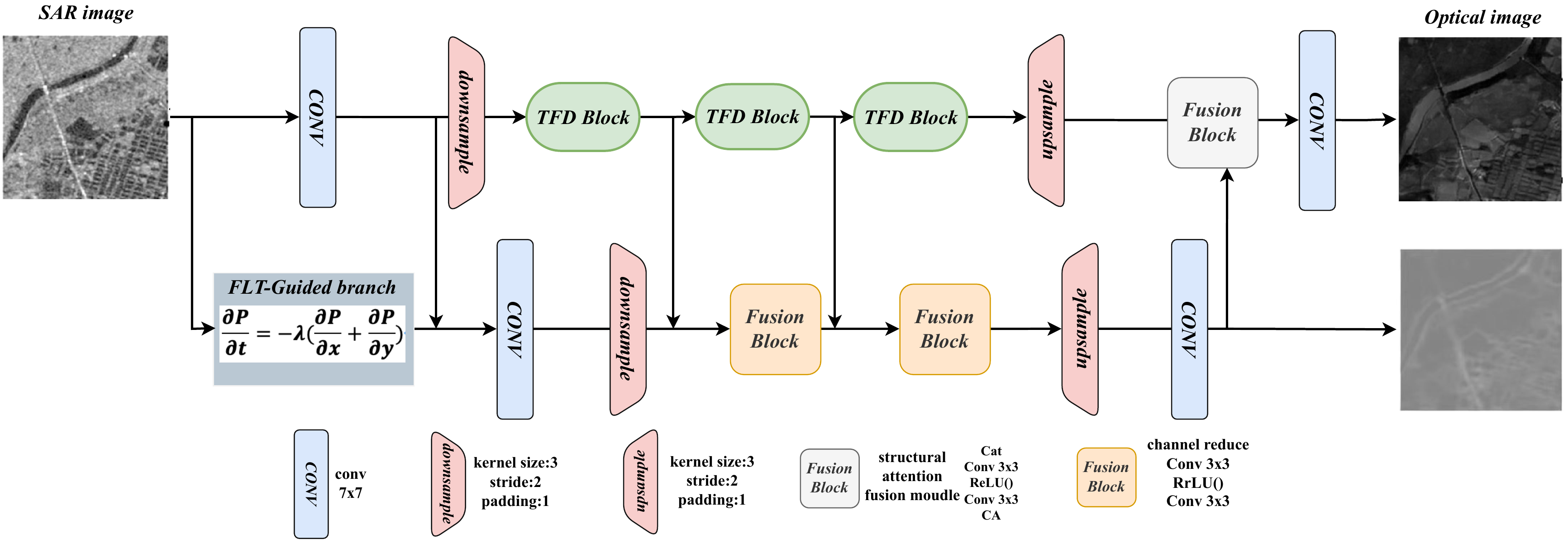}
	\caption{The generator structure of the proposed S2O-TDN, which is in line with CycleGAN. The generator consists of a backbone with three TFD residual blocks (Section \ref{subsec:tfd}) inserted and a parallel FLT-guided branch (Section\ref{subsec:flt}) for further feature interactions. The high-dimensional tensor information output from the FLT-guided branches is output as a one-channel two-dimensional tensor by the convolution layer for intuitive visual understanding.} 
	\label{overall}
\end{figure*}

Based on the above motivation, we go a step further toward solving the S2O problem from the perspective of the evolution of SAR images and make the first attempt to apply the thermodynamic theory to the S2O model design. Specifically, we propose a Thermodynamics-inspired Network (S2O-TDN) for S2O image translation in this paper. First, we construct a basic third-order finite difference (TFD) residual block inspired by the TFD equation of thermodynamics, which is utilized to construct our backbone network. The helpful information is extracted from the SAR images by accumulating and strengthening features at different layers.
Meanwhile, to address the pixel diffusion problem in the S2O image translation process, an FLT-guided branch is designed explicitly based on the first law of thermodynamics (FLT). It facilitates the transition of feature values from the unstable diffusion state to the stable state and aims to regularize the feature diffusion and maintain the image structures in the S2O image translation process. Putting them together in the GAN framework, we get our S2O-TDN as shown in Fig.~\ref{overall}. Experiments on the public SEN1-2 benchmark~\cite{guo2021edge} illustrate the advantages of the proposed S2O-TDN over the most advanced methods in terms of objective indications and visual quality.
.

The key contributions of this paper are summarized in three areas.
\begin{itemize}
\item A thermodynamic perspective is taken on the S2O image translation task, and accordingly, we propose a novel S2O-TDN that follows a clear and interpretable design principle. This is the first time that thermodynamic theories are brought into S2O image translation networks.
\item Inspired by the third order finite difference equation (TFD), \textit{i.e.}, the TFD residual block is used to build the backbone network. And motivated by the first law of thermodynamics (FLT), \textit{i.e.}, the FLT-guided branch is developed. They help the proposed S2O-TDN to learn a better nonlinear mapping between inter-domain features for S2O image translation while preserving the geometric structures by mitigating the pixel diffusion problem.
\item The proposed S2O-TDN model is experimentally tested on the currently popular SEN1-2 dataset with better objective metrics and visual image quality. Generated optical images with finer structure and better geometry.
\end{itemize}

Our methodology performs well in the S2O image transformation process and can also be used for general image transformations. The general framework of this paper is as follows. Section II contains several concepts and theories that are discussed in this paper. Section III explains the overall structure of the S2O-TDN network. Section IV provides comparative tests and analysis of the network results of this paper. Section V gives a summary of the major contributions of the paper.

\section{Related Work}

\subsection{SAR-to-Optical Image Translation}
Image-to-image translation has evolved as one of the most important research subjects in deep learning~\cite{9770477,8913489}. Initially, only content loss functions was leveraged to calculate the loss, like L1-normal loss or L2-normal loss, which resulted in poor image quality~\cite{zhang2019deep,zhang2019neural}. Since GANs can generate images with good visual properties, many scholars have made extensive attempts to apply generative adversarial networks to inter-image translation tasks \cite{9527389}.
Among the S2O image translation methods based on deep learning, the GAN-based approach shows promising performance because its generators can produce delicate visual images. A series of methods such as Feature-Guiding Generative Adversarial Networks (FGGAN)~\cite{2020Feature}, Edge-Preserving Convolutional Generative Adversarial Networks (EPCGAN)~\cite{guo2021edge}, and Supervised Cycle-consistent Generative Adversarial Networks (S-CycleGAN)~\cite{wang2019sar}, have been proposed. By leveraging the provided information such as class labels in the generator of the original GAN, Conditional Generative Adversarial Networks (cGAN)~\cite{2021SAR} can learn an adaptive sample generation process for different classes during training, thus enabling the model to generate more realistic images belonging to a given class. However, for the S2O image conversion task, the generated optical images usually do not have the desired fine structure since there is a huge inter-domain gap existing between the SAR image and the optical image. Isola \textit{et al.} presented a pix2pix~\cite{isola2017image} model based on cGAN and utilized the input image as a condition for image translation and learns the relationship of mapping between the input and output images for generating a specific output image. But the model requires known image pairs, which are often difficult to obtain in the real world. Zhu \textit{et al.} presented the Cycle-Consistent Adversarial Networks (CycleGAN)~\cite{zhu2017unpaired}, which utilized two mirror-symmetric GANs under the cyclic consistency loss to form a ring network, \textit{i.e.}, a structure that does not require an input pair of images. However, the generated images by the above GAN-based methods do not perform well on the S2O image translation task by suffering from geometric distortion, since there is a huge domain gap between the SAR and optical images, and there are no explicit constraints on the evolution of pixels from SAR images to optical images, making them difficult to learn an effective mapping function between the inter-domain features.

In contrast to the above approach, we build our S2O image translation model by taking the explicit design guideline of thermodynamic theory. Specifically, from the microscopic perspective, the evolution of pixels during the S2O image translation process can be analogized to the particles of thermodynamics. Accordingly, we devise a basic TFD residual block and an FLT-guided branch based on the third-order finite difference equation and the first law of thermodynamics. They constrain the movement of the pixels and mitigate the challenges in the S2O image translation task such as pixel diffusion, delivering better results.

\subsection{Neural Partial Differential Equations}
Deep learning has reshaped many research fields in computer vision and made significant progress in recent years~\cite{2017Compositional,8463611}. To mitigate the degeneration problem of the representation ability of deep neural networks built by stacking many plain convolutional layers, He \textit{et al.} \cite{2016Deep} presented a new residual network structure (ResNet) to replace a single plain convolution layer, which can be easily scaled up to more than 1,000 layers and enhance the network capacity drastically. The deep residual learning idea has been used in many vision tasks~\cite{Ref1,Ref2,Ref3}, especially in low-level image processing tasks. For instance, Zhang \textit{et al.} proposed the DnCNNs for image super-resolution by introducing the residual learning idea into the feed-forward convolutional neural networks, where a residual map between the input image and ground truth target is learned. Zhang \textit{et al.} developed a lightweight fully point-wise dehazing network with residual connections to learn multi-scale haze-relevant features in an end-to-end way, delivering good dehazing results in a fast inference speed~\cite{zhang2019famed}. E \cite {Weinan2017A} explored the potential possibility that ordinary differential equations (ODE) can be used in the design of neural networks and interpreted the residual network as a discrete dynamical system. Recently, many ODE-inspired networks have been proposed~\cite{lu2018beyond,he2019ode}. For example, He \textit{et al.} \cite{he2019ode} investigated the forward Euler method of dynamical systems with the similarity of residual structures and proposed a novel network inspired by ODE for image super-resolution.

We also devise our S2O image translation model based on neural ODE but interpret the S2O image translation process as a specific dynamical system, \textit{i.e.}, thermodynamics. From the macro perspective, we can regard the layers of neural networks in the S2O image translation process as the status forward units of time in a thermodynamic system. Since, according to the first law of thermodynamics, the thermal field tends to transfer from an unstable high-temperature state to a more stable low-temperature state~\cite{romano2010variational}, we consider following the first law of thermodynamics and designing a guided branch to regularize the feature diffusion. In addition, we also leverage the third-order finite difference equation of thermodynamics to replace the first-order equation of the existing dynamical system for designing a TFD residual structure.

\section{Method}
\subsection{Preliminary}
From the view of thermodynamics, the particles in the closed heat field always vibrate, collide and move under the constraint of temperature difference. As for the S2O image translation process, the pixels in the image are spreading and moving all the time under the constraints of the loss function and the optimization function. These pixel movements tend to cause a re-organization of the image information so that the image information in the SAR domain transforms into the optical domain and finally the network reaches the stable state after learning the desired image mapping function. Based on this analogy, we can draw inspiration from the thermal field motion theory for the guidance of the S2O image conversion process. 

Based on the thermodynamic theory, we propose a novel Thermodynamics inspired Network (S2O-TDN) for S2O image translation. The generator of the proposed S2O-TDN mainly contains a third-order finite difference (TFD) residual structure and the first law of thermodynamics (FLT) guided branch. Specifically, we explore the similarity between thermodynamic particle flow and pixels flow of the S2O image translation process. And we utilize the finite difference theory of thermodynamics to construct the TFD residual structure so that the residual network can fully integrate inter-domain information and learn a desired nonlinear mapping. Meanwhile, considering the pixel diffusion problem in the S2O image translation process, we design the FLT-guided branch according to FLT, which makes the network internally converge from a pixel non-stationary diffusion state to a stable state to preserve image details.

\begin{figure}
\centering
\includegraphics[width=1.0\columnwidth]{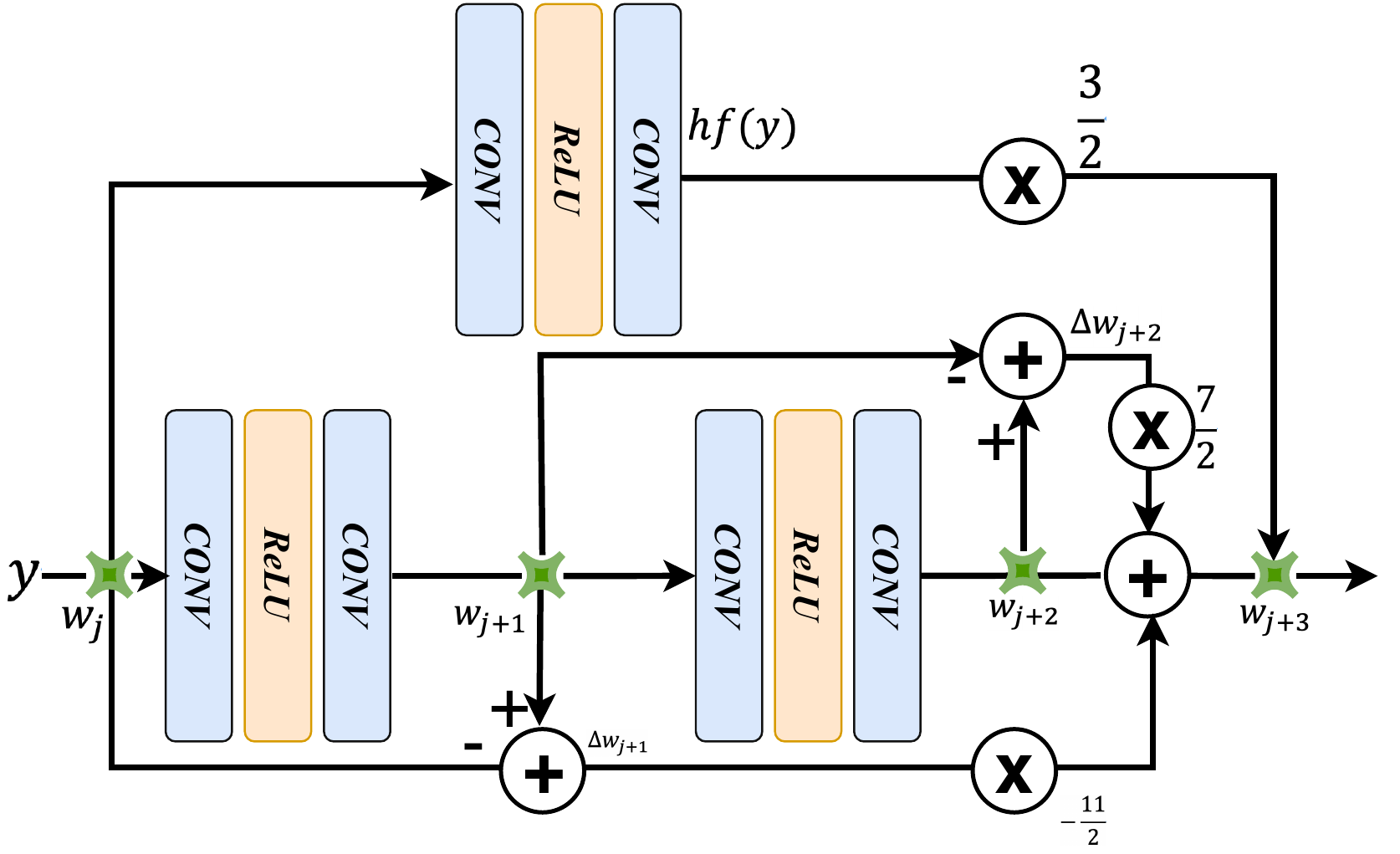}  
\caption{Structure of the TFD residual block.}
	\label{2}
 \vspace{-4mm}
\end{figure}

\subsection{Third-order Finite Difference Residual Structure}
\label{subsec:tfd}
Since there exist a large domain gap between SAR and optical images, it is difficult to achieve good nonlinear mapping in the S2O image translation process. Accordingly, we put forward a residual-specific block to achieve better mapping of features, so that various features in the SAR image domain can be encoded to map to the optical image domain and generate superior quality optical images. Thus, we apply the idea of third-order finite difference equations in thermodynamics to construct the third-order finite difference (TFD) residual block.

Specifically, we use the finite difference method to discretize the ODE and build a novel SR residual network, TFD block. Since the third-order derivatives in the finite difference equation can extract multidimensional information during the interaction, we utilize the third-order finite difference to devise a residual network structure. The third-order finite difference is defined as:

\begin{equation}\label{eq:TFD}
    \left( \frac{\partial w}{\partial y} \right) _j=\frac{-11w_{j}+18w_{j+1}-9w_{j+2}+2w_{j+3}}{3\Delta y},
\end{equation}
where $w$ represents an objective that is dependent on the input $y$. $\Delta y$ refers to the difference between the input and output of the TFD residual block. The above equation can be reformulated as follows.

\begin{equation}\label{eq:TFD2}
    \left( \frac{\partial w}{\partial y} \right) _j\Delta y=-\frac{11}{3}w_{j}+6w_{j+1}-3w_{j+2}+\frac{2}{3}w_{j+3}.
\end{equation}
A change of the target $w$ from layer $j$ to layer $j+3$ is exhibited on the left of the above Eq.~\eqref{eq:TFD2} and is approximated in the neural network by a neural module approved as $h f\left( y\right)$. The above equation can be further derived as:

\begin{align}\label{eq:TFD_residual}
h f(y) &=\frac{11}{3}\left(w_{j+1}-w_{j}\right)-\frac{7}{3}\left(w_{j+2}-w_{j+1}\right) \nonumber \\&-\frac{2}{3}w_{j+2}+\frac{2}{3}w_{j+3}.
\end{align}%
To simplify the above equation, we use $\Delta w_{j+2}$ and $\Delta w_{j+1}$ to represent the residuals over $w_{j+2}$ and $w_{j+1}$ and the residuals over $w_{j+1}$ and $w_{j}$, respectively. Then, we can obtain the final mathematical form of the TFD residual block.

\begin{equation}\label{eq:TFD_residual_block}
  w_{j+3}=\,\,w_{j+2}+ \frac{7}{2}\Delta w_{j+2}-\frac{11}{2}\Delta w_{j+1}+\frac{3}{2}hf(y).
\end{equation}
In this paper, a TFD residual block is designed to implement Eq.~\eqref{eq:TFD_residual_block}. Specifically, we leverage a sequence of convolutional layers with a ReLU layer for designing ${h}f(y)$. Figure~\ref{2} depicts the detailed structure of the TFD residual block.

\begin{figure*}[t!]
	\centering
	\includegraphics[scale=1.0]{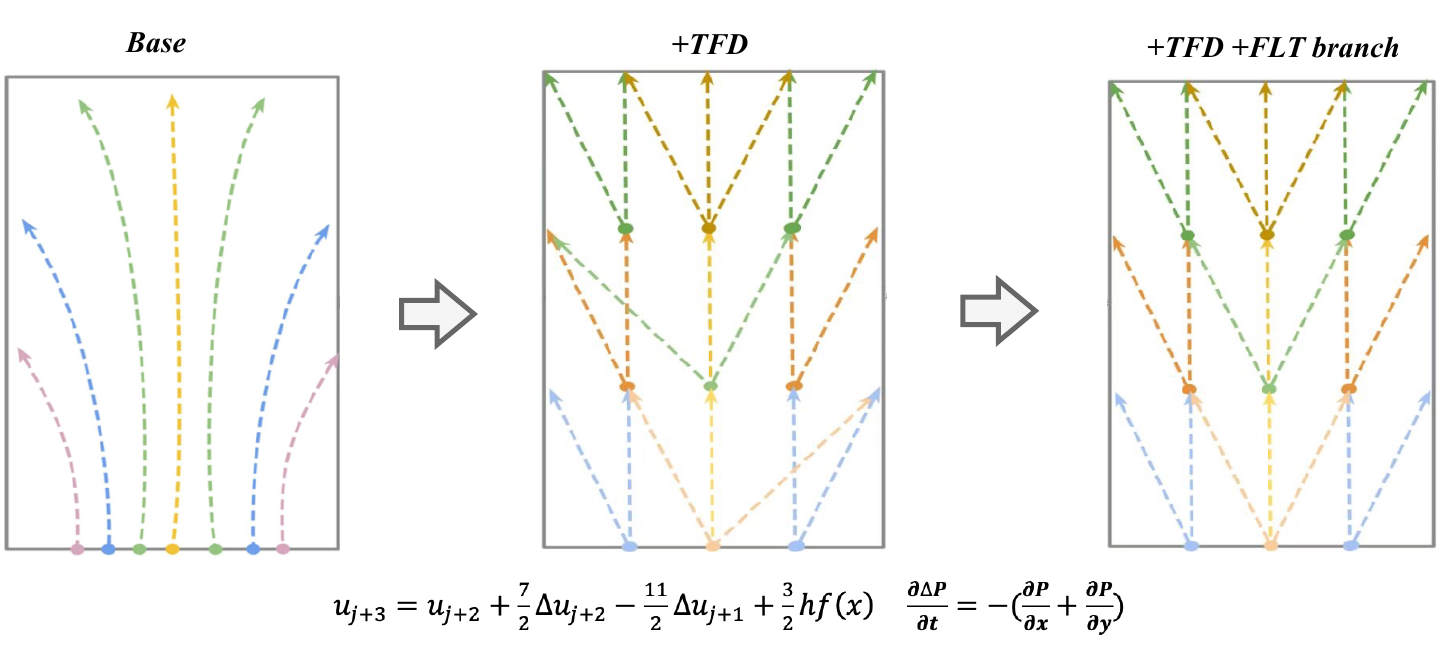}
	\caption{Influence of the TFD residual structure and FLT-guided branch on pixel flow during S2O image translation.}
	\label{pic_2}
\end{figure*}

\subsection{First Law of Thermodynamics Guided Branch}
\label{subsec:flt}

As shown in Figure~\ref{pic_2}, the uncertainty of pixel motion during S2O leads to irregular particle forward motion, and this forward motion needs to be constrained to suppress the possibility of pixels being dispersed to some extent. To solve this problem, we designed the FLT bootstrap branch to obtain discrete-time particle values by approximating the continuous deviation using a thermodynamic TFD. 

Figure 4 shows, from left to right, the simulated trajectories of irregular particles moving forward after being constrained by the TFD residual block and the FLT guided branch, respectively.

\subsubsection{The First Law of Thermodynamics}
In the thermodynamic field, the energy of a system is conserved, which is known as the FLT. It points out that all matter in nature has energy, which can neither be created nor destroyed. But the energy can be transferred from one to another. And the total amount of energy remains the same.
In other words, after the thermodynamic system reaches the final state from the initial state through any process, the increment of internal energy should be equal to the difference between the heat transferred to the system by the outside world and  the energy work produced by the system on the outside, which can be quantified by the following mathematical formula:

\begin{equation}
Q=\triangle U+A,
\end{equation}
where $Q$ is the heat gained by the system per unit of time. $\triangle U$ represents the energy added to the system per unit of time. $A$ represents the power done by the system to the outside. The physical meaning of this formula is that the change in total energy in a system is equal to the energy difference between the outlet and the entrance plus the change in energy within the system.

Similarly, treating the S2O image translation network as a dynamic system, the difference between input and output pixels ${P_f}$ plus the change of pixel values in the training image itself ${P_v}$ equals the overall variation of pixel values throughout the network ${\triangle P}$, which can be described as.

\begin{align}\label{3}
\triangle P=P_f+P_v.
\end{align}%

\subsubsection{The variance of the input and output pixel values $P_f$}
During the S2O image translation, we assume that an image can be arbitrarily divided into several small regions, each of which we label as $\Omega$. $\Omega$ is small enough. Each $\Omega$ pixel per unit time is $P$, and its pixel-value density with partial differentiation in the $x$-axis and $y$-axis directions are $p_{x}$ and $p_{y}$, respectively. Then we can derive the relationship between the pixel value and pixel value density of the image in each $\Omega$ region in one unit of time as,

\begin{equation}\label{4}
dP_{x}=p_{x}dydt,
\end{equation}
\begin{equation} 
dP_{y}=p_{y}dxdt,
\end{equation}
where $P_{x}$ and $P_{y}$ denote the pixel values of $x$-axis and $y$-axis directions per unit time at the small region of the input image $\Omega_{in}$, respectively. $p_{x}$ and $p_{y}$ are the pixel density values of $x$-axis and $y$-axis directions per unit time at the small region of the input image $\Omega_{in}$, respectively. Given the same reasoning process in the two directions: $x$-axis and $y$-axis, we only elaborate on the $x$-axis related process, and the $y$-axis is the same. We assume that $P_{x+dx}$ and $p_{x+dx}$ represent the pixel values and the pixel density values of $x$-axis direction per unit time at the small region of the output image $\Omega_{out}$. In the same way as Eq.~\eqref{4}, we can obtain the relation of the output $P_{x+dx}$ and $p_{x+dx}$ as:

\begin{equation}\label{6}
dP_{x+dx}=p_{x+dx}dydt.
\end{equation}

According to the relationship between difference and derivative, Eq.~\eqref{4} and Eq.~\eqref{6} can be further obtained as follows:

\begin{equation}
p_{x}-p_{x+dx}=-\frac{\partial p_{x}}{\partial x}dx.
\end{equation}
Then the difference between the $\Omega_{in}$ and $\Omega_{out}$ pixel value on the $x$-axis is:

\begin{equation}
dP_{x}-dP_{x+dx}=-\frac{\partial p_{x}}{\partial x}dxdydt.
\end{equation}
After adding the pixel differences of the $x$-axis and $y$-axis directions, we obtain the pixel differences of each $\Omega_{in}$ and $\Omega_{out}$ in unit time:

\begin{align}
 dP_{x}-dP_{x+dx}+dP_{y}-dP_{y+dy}=-\left( \frac{\partial p_{x}}{\partial x}+\frac{\partial p_{y}}{\partial y}\right) dxdydt.
\end{align}%

Finally, we add up the difference between the pixel values of $\Omega_{in}$ and $\Omega_{out}$ for all small regions of $\Omega$. In this way, we can obtain the difference ${P_f}$ between the input and output pixels of the whole image during the S2O image translation process, which is caused by pixel diffusion. When each $\Omega$ tends to infinity, the above summation process can be approximated as the integration of a two-dimensional image:

\begin{align}\label{10}
    P_f&=\iint (dP_{x}-dP_{x+dx}+dP_{y}-dP_{y+dy})dxdy \nonumber\\
    &=\iint [-\left( \frac{\partial p_{x}}{\partial x}+\frac{\partial p_{y}}{\partial y}\right) dxdydt]dxdy \nonumber\\
    & = -\left( \frac{\partial p_{x}}{\partial x}+\frac{\partial p_{y}}{\partial y}\right)dt.
\end{align}%

\subsubsection{The change of the image's own pixel value $P_v$}
Besides the variations due to the pixel inflow and outflow, the decrease and increase of the pixel values of the training image itself will also impact the global values. The pixel value will change over time for every point in the microcluster.

\begin{equation}\label{11}
dP_{v}= p_{v}dxdydt,
\end{equation}
where $p_{v}$ represents the pixel value per unit area of the image per unit of time, which is the so-called pixel density. $P_{v}$ is the increment of pixel values inside the image.

\subsubsection{The overall changes $\triangle P$}
Combining Eq.~\eqref{3}, Eq.~\eqref{10} and Eq.~\eqref{11}, $\triangle P$ can be obtained. 

\begin{align}\label{12}
\triangle P& = -\left( \frac{\partial p_{x}}{\partial x}+\frac{\partial p_{y}}{\partial y}\right)dt+ p_{v}dxdydt.
\end{align}%

In this paper, considering that the pixel value of each point in the trained image is fixed, the value of the pixel itself does not change, \textit{i.e.}, $p_{v}=0$. Therefore we only focus on the incremental pixel values of the image input and output. When $\Omega$ tends to infinity, the pixel density $p_{x}$, $p_{y}$ within each $\Omega$ can be approximated as the pixel value $P$ in that region,  Therefore, Eq.~\eqref{12} can be rewritten:

\begin{equation}\label{14}
\frac{\partial \triangle P}{\partial t} =-\left(\frac{\partial P}{\partial x}+\frac{\partial P}{\partial y}\right).
\end{equation}
That is, in the time domain, the change rate of pixel difference $\frac{\partial \triangle P}{\partial t}$ and the partial derivative of a pixel in the $x$ and $y$ axes $\frac{\partial P}{\partial x}$ and $\frac{\partial P}{\partial y}$ can be obtained during the S2O image translation process. Therefore, based on the similarity between the laws of motion of particles in thermodynamics  and the pixel flow motion during S2O image translation, we can devise our FLT-guided branch accordingly.

As shown in Eq.~\eqref{14}, we use some convolution layers with fixed kernels to realize the function of the right side of the equation. Specifically, using convolution kernels like $W_{x}=[0,1,0;\mathbf{0};0,-1,0]$, $W_{y}=[\mathbf{0};-1,0,1;\mathbf{0}]$, we can obtain the derivative of the image along the $x$ and $y$ axes. The convolution layers have constant kernels, which we name the FLT-guided heads. In addition to utilizing fixation convolution kernel learning. Several fusion blocks were also designed to facilitate functional interaction between the spine and the FLT-guided head and to further enhance the detailed features. The FLT-guided head structure information is available in Figure~\ref{overall}.
The output feature maps of the FLT-guided head are processed by a simple convolution and downsampling process, and then the outputs of their TFD residual blocks are fused using two consecutive fusion modules. Each fusion module is composed of three convolutions. Finally, a side scan, i.e., the FLT-guided image, is obtained following an upsampling layer and a prediction layer.
The high-dimensional tensor information output from the FLT-guided branches is output as a one-channel two-dimensional tensor by the convolution layer for intuitive visual understanding.

\subsection{Overall Structure of S2O-TDN}

We design the S2O-TDN model built on the CycleGAN architecture. The backbone network of the generator is composed of three TFD residual blocks (Section \ref{subsec:tfd}) and a parallel FLT-guided branch (Section \ref{subsec:flt}), the exact structure of which is described in detail in Figure ~\ref{overall}. In the backbone, SAR images are first fed into convolutional and downsampling layers and then go through three TFD residual blocks for feature aggregation and inter-domain feature transformation. The parallel FLT-guided branch aims to constrain pixel diffusion in the S2O image translation process. To be specific, we first feed SAR images to the head of the FLT-guided branch to normalize feature spread and maintain image structure during S2O image conversion. In the fusion block, we fuse the output feature with the feature of the remaining TFD blocks. Then after simply going through the upsampling and convolution layers, we obtain a secondary output: an FLT-guided image. Finally. we feed the FLT-guided image into the backbone fusion block of facilitate feature interactions and drive the generator to produce the final fine-grained optical image. A specific loss (Section~\ref{subsec:loss}) is utilized to optimize  the proposed S2O-TDN.

\subsection{Loss Function}
\label{subsec:loss}
In the proposed SO-TDN, we design a TD loss $\mathcal{L} _{TD}$ to highlight the high frequency of details. $\mathcal{L} _{TD}$ can be defined as:

\begin{align}
    \mathcal{L} _{TD}&=\mathcal{L} _{TFD} + \mathcal{L} _{FLT-guided} \nonumber\\
    &=\mathbb{E} _{S,O\sim p\left( S,O \right)}\left\| \phi \left( O \right) -\phi \left( G_{TFD}\left( S \right) \right) \right\| + \nonumber\\
    &+\mathbb{E} _{S,O\sim p\left( S,O \right)}\left\| \phi \left( O \right) -G_{FLT-guided}\left( S \right) \right\|  ,
\end{align}%
where $\mathcal{L} _{FLT-guided}$ and $\mathcal{L} _{TFD}$ are the lateral output of the neural module used to supervise the FLT-guided and the loss of the final forecast of the backbone network, separately.  $\phi \left( \cdot \right)$ denotes the FLT-guided head function, as discussed in (Section \ref{subsec:tfd}). In the above equation, $S$ is used to represent the real SAR image. The optical image is represented using $O$. In the limit of the loss function, higher frequency features $\phi \left( O \right)$ obtained in the true optical map are compatible with the higher frequency features $\phi \left( G_{TFD}\left( S \right) \right)$  of the output optical images and the FLT-guided images generated by the FLT-guided neural module $G_{FLT-guided}\left( S \right)$ are compatible. The  $\mathcal{L} _{FLT-guided}$ then becomes part of the FLT-guided neural module.

Then we combine the proposed Thermodynamics (TD) loss $\mathcal{L} _{TD}$ with several loss functions including the adversarial loss $\mathcal{L} _{GAN}$, cycle consistency loss $\mathcal{L} _{cyc}$, pixel loss $\mathcal{L} _{pix}$, and perceptual loss $\mathcal{L} _{per}$. These losses have been commonly used in previous works \cite{isola2017image,zhu2017unpaired,2021SAR} and show their effectiveness in supervising the training of S2O image translation networks. This leads to the definition of the overall loss function as follows:

\begin{align}
  \mathcal{L} &=\mathcal{L} _{GAN}+\lambda _{pix}\mathcal{L} _{pix}+\lambda _{per}\mathcal{L} _{per}+\lambda _{cyc}\mathcal{L} _{cyc}+\nonumber\\
&+\lambda _{TD}\mathcal{L} _{TD},
\end{align}
where $\lambda _{pix}$, $\lambda _{per}$, $\lambda _{cyc}$ and $\lambda_{TD}$ are the coefficient of the loss function: $\mathcal{L} _{GAN}$, $\mathcal{L} _{cyc}$, $\mathcal{L} _{pix}$, and $\mathcal{L} _{per}$, respectively. And we empirically set the above hyper-parameters to 10 in our experiments.

\section{Experimental Results and Analysis}
\begin{figure*}[t!]
	\centering
	\includegraphics[width=1.0\linewidth]{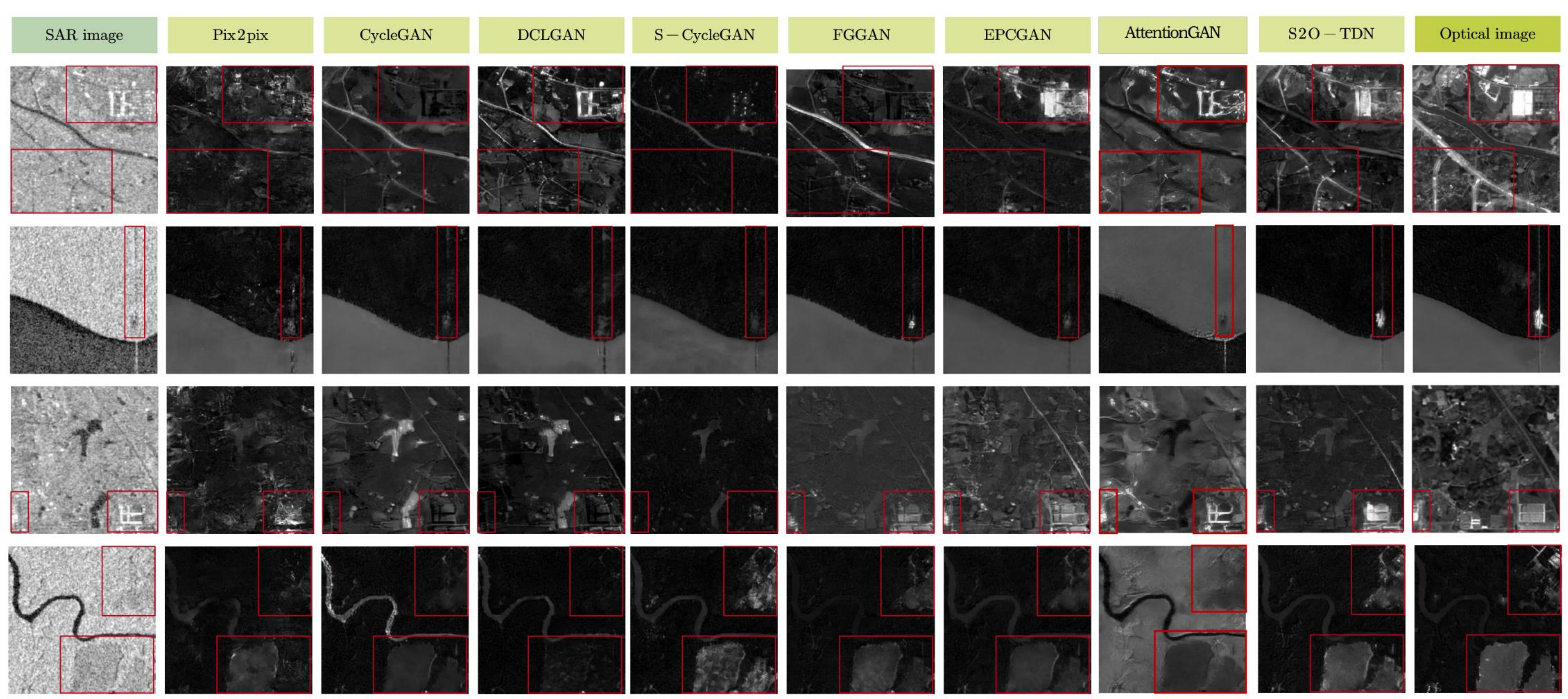}
	\caption{The visual results of the proposed S2O-TDN with the other S2O image translation methods including Pix2pix~\cite{isola2017image}, CycleGAN~\cite{wang2019sar}, DCLGAN~\cite{2021Dual} S-CycleGAN~\cite{zhu2017unpaired}, FGGAN~\cite{2020Feature}, EPCGAN~\cite{guo2021edge}, AttentionGAN~\cite{2019AttentionGAN}.}
	\label{fig:visual_results}
\end{figure*}

\begin{figure*}[t]
	\centering
	\includegraphics[width=1.0\linewidth]{Result02.pdf}
	\caption{Structure information comparison of the proposed S2O-TDN and the other S2O image translation methods including Pix2pix~\cite{isola2017image}, CycleGAN~\cite{wang2019sar}, DCLGAN~\cite{2021Dual} S-CycleGAN~\cite{zhu2017unpaired}, FGGAN~\cite{2020Feature}, EPCGAN~\cite{guo2021edge}, AttentionGAN~\cite{2019AttentionGAN}.}
	\label{fig:visual_structure}
 \vspace{-4mm}
\end{figure*}

\subsection{Datasets}
The SEN1-2 dataset contains 282,384 pairs of image blocks within a series of paired SAR images and optical images. We crop the average distance of all images involved in the training to $256\times256$~\cite{guo2021edge}.
A popular approach to choosing a dataset is to extract certain image batches from the images and use them as the training set and the remaining as the test dataset while assuming that no pixel overlaps between every two images.
This approach makes sense when pairwise data sources for a particular problem are difficult to access. Nevertheless, there is a great similarity between small image blocks cropped on the same large image. By training the network on the same image blocks that the test set has, the performance of the model on the test set will be better than it should be, and the models' robustness cannot be reflected in the results of such experiments.

We obtain 1600 pairs of high-resolution optical SAR images from the SEN1-2 database as the training dataset and 300 pairs of test database (denoted by Test1) with the size of 256$\times$256~\cite{guo2021edge}. These images are clipped from the initial 258 pairs of high-resolution optical-SAR images. The test and the train datasets cover various topographic classes, such as forests, lakes, mountains, rivers, buildings, farmlands, and roads. In addition, 52 pairs of complex mountain scenes and 68 pairs of complex suburban scenes are selected as two other test sets named Test2 and Test3. The scenes in Test 2 and Test 3 do not appear in the learning process and are used as unseen data to effectively evaluate the robustness of the proposed S2O-TDN.

\subsection{Implementation Details}
In this paper, we employ the ADAM optimizer, where  $\beta _1=0.5$, $\beta _2=0.999$. Our model is trained over 200 epochs, the batch size is set to 1, the learning rate is equal to $2\times 10^{-4}$, and decreases linearly from the 100th epoch until it reaches 0. We select Pix2pix~\cite{isola2017image}, CycleGAN~\cite{wang2019sar}, DCLGAN~\cite{2021Dual} S-CycleGAN~\cite{zhu2017unpaired}, FGGAN~\cite{2020Feature}, EPCGAN~\cite{guo2021edge}, AttentionGAN~\cite{2019AttentionGAN} as the representative S2O image translation methods for comparison. CycleGAN and DCLGAN are unsupervised methods, and the others are supervised methods. Pix2pix, CycleGAN, and DCLGAN are architectures for generic images and the others are designed specifically for the S2O image translation process. 
We train the model by using Pytorch on an NVIDIA GTX 2080Ti GPU. 

\begin{table*}
\normalsize
\centering
\setlength{\tabcolsep}{6mm}
\caption{Quantitative results of different methods. 
}
\begin{tabular}{ccccccc}
\hline
{\multirow{2}{*}{Method}} & \multicolumn{3}{c}{PSNR}  &  \multicolumn{3}{c} {SSIM}                                                                                          \\ \cline{2-7} 
\multicolumn{1}{c}{}                        & \multicolumn{1}{c}{Test1}        & \multicolumn{1}{c}{Test2}        & \multicolumn{1}{c}{Test3}        & \multicolumn{1}{c}{Test1}         & \multicolumn{1}{c}{Test2}         & \multicolumn{1}{c}{Test3}         \\ \hline
\multicolumn{1}{c}{Pix2pix}                 & \multicolumn{1}{c}{17.18}          & \multicolumn{1}{c}{15.93}          & \multicolumn{1}{c}{16.46}          & \multicolumn{1}{c}{0.3452}          & \multicolumn{1}{c}{0.2664}          & \multicolumn{1}{c}{0.2715}          \\ \hline
\multicolumn{1}{c}{CycleGAN}                & \multicolumn{1}{c}{16.51}          & \multicolumn{1}{c}{15.13}          & \multicolumn{1}{c}{15.91}          & \multicolumn{1}{c}{0.3422}          & \multicolumn{1}{c}{0.2956}          & \multicolumn{1}{c}{0.2896}          \\ \hline
\multicolumn{1}{c}{S-CycleGAN}              & \multicolumn{1}{c}{18.05}          & \multicolumn{1}{c}{15.67}          & \multicolumn{1}{c}{16.53}          & \multicolumn{1}{c}{0.4082}          & \multicolumn{1}{c}{0.2899}          & \multicolumn{1}{c}{0.2841}          \\ \hline
\multicolumn{1}{c}{FGGAN}                   & \multicolumn{1}{c}{18.56}          & \multicolumn{1}{c}{16.80}          & \multicolumn{1}{c}{17.14}          & \multicolumn{1}{c}{0.4438}          & \multicolumn{1}{c}{0.3625}          & \multicolumn{1}{c}{0.3302}          \\ \hline
\multicolumn{1}{c}{EPCGAN}                  & \multicolumn{1}{c}{18.98}          & \multicolumn{1}{c}{16.54}          & \multicolumn{1}{c}{17.46}          & \multicolumn{1}{c}{0.4491}          & \multicolumn{1}{c}{0.3615}          & \multicolumn{1}{c}{0.3454}          \\ \hline
\multicolumn{1}{c}{AttentionGAN}                 & \multicolumn{1}{c}{12.50}          & \multicolumn{1}{c}{14.70}          & \multicolumn{1}{c}{15.18}          & \multicolumn{1}{c}{0.3239}          & \multicolumn{1}{c}{0.1914}          & \multicolumn{1}{c}{0.2374}          \\ \hline
\multicolumn{1}{c}{\textbf{S2O-TDN}}           & \multicolumn{1}{c}{\textbf{19.16}} & \multicolumn{1}{c}{\textbf{17.93}} & \multicolumn{1}{c}{\textbf{18.26}} & \multicolumn{1}{c}{\textbf{0.4736}} & \multicolumn{1}{c}{\textbf{0.4053}} & \multicolumn{1}{c}{\textbf{0.3705}} \\ \hline
\end{tabular}
\label{tab:quantitative}
\end{table*}

\subsection{Quantitative Results}
We utilize signal-to-noise ratio (PSNR), mean square error (MSE), and structural similarity (SSIM)~\cite{wang2004image} as benchmarks to objectively quantify the proposed S2O-TDN and the existing translation approaches. MSE indicates the mean difference between the corresponding pixels. The pixel values of the image are normalized from the range of 0-255. Its MSE index is then calculated, indicating that the smaller the MSE, the image has lower the level of distortion. The MSE of the average difference between the optical image produced by the S2O translation network and the real optical image is called PSNR. The PSNR index is inversely correlated with image distortion.
Although the PSNR metric is accurate enough from an objective point of view, the visual presentation does not show the same results. Therefore, SSIM is used to evaluate the similarity of luminance, contrast, and texture. The PSNR metric is inversely correlated with image distortion.

We present the PSNR and SSIM ~\cite{wang2004image} results of the different methods on three test datasets. The results of different methods on three test sets are listed in Table~\ref{tab:quantitative}. It is evident that the proposed S2O-TDN obtains the best performance. For example, in Test 1, the EPCGAN is 0.2 dB and 0.0245 SSIM higher than the second-best EPCGAN. It is worth noting that our method has a significant performance advantage on Test 2 and Test 3, which demonstrates the excellent robustness of our method. Since the proposed S2O-TDN effectively extracts multi-level features through residual learning in TFD residual blocks, and uses FLT-guided neural modules to extract and reconstruct structural information to further assist image transformation.
The FLT-guided branch has well-defined designs based on FLT functions that facilitate the generation of target optical images. However, the image translation networks, such as CycleGAN and Pix2pix, do not explicitly model the structure and fail to constrain the reconstruction process of pixels, thus their effectiveness is limited. And the existing CNN-based S2O translation methods pay less attention to structure guidance and feature extraction for S2O translation through the design of the network structure, resulting in poor performance. 

\begin{table*}
\normalsize
\centering
\setlength{\tabcolsep}{6mm}
\caption{Ablation study results.}
\begin{tabular}{ccccccc}
\hline
\multicolumn{1}{c}{\multirow{2}{*}{Method}} & \multicolumn{3}{c}{PSNR}                                                                                       & \multicolumn{3}{c}{SSIM}                                                                                          \\ \cline{2-7} 
\multicolumn{1}{c}{}                        & \multicolumn{1}{c}{Test1}        & \multicolumn{1}{c}{Test2}        & \multicolumn{1}{c}{Test3}        & \multicolumn{1}{c}{Test1}         & \multicolumn{1}{c}{Test2}         & \multicolumn{1}{c}{Test3}         \\ \hline
\multicolumn{1}{c}{Base}                 & \multicolumn{1}{c}{18.19}          & \multicolumn{1}{c}{16.99}          & \multicolumn{1}{c}{16.96}          & \multicolumn{1}{c}{0.4351}          & \multicolumn{1}{c}{0.3696}          & \multicolumn{1}{c}{0.3329}          \\ \hline
\multicolumn{1}{c}{+FLT-guided}                & \multicolumn{1}{c}{18.61}          & \multicolumn{1}{c}{17.29}          & \multicolumn{1}{c}{17.53}          & \multicolumn{1}{c}{0.4529}          & \multicolumn{1}{c}{0.3702}          & \multicolumn{1}{c}{0.3480}          \\ \hline
\multicolumn{1}{c}{+TFD}              & \multicolumn{1}{c}{18.59}          & \multicolumn{1}{c}{17.60}          & \multicolumn{1}{c}{17.96}          & \multicolumn{1}{c}{0.4441}          & \multicolumn{1}{c}{0.3825}          & \multicolumn{1}{c}{0.3596}          \\ \hline
\multicolumn{1}{c}{+Poly2 +FLT-guided}              & \multicolumn{1}{c}{18.78}          & \multicolumn{1}{c}{17.87}          & \multicolumn{1}{c}{17.33}          & \multicolumn{1}{c}{0.4644}          & \multicolumn{1}{c}{0.3986}          & \multicolumn{1}{c}{0.3541}          \\ \hline
\multicolumn{1}{c}{+RK2 +FLT-guided}              & \multicolumn{1}{c}{18.54}          & \multicolumn{1}{c}{17.12}          & \multicolumn{1}{c}{16.69}          & \multicolumn{1}{c}{0.4408}          & \multicolumn{1}{c}{0.3814}          & \multicolumn{1}{c}{0.3329}          \\ \hline
\multicolumn{1}{c}{\textbf{+TFD +FLT-guided}}                   & \multicolumn{1}{c}{\textbf{19.16}}          & \multicolumn{1}{c}{\textbf{17.94}}          & \multicolumn{1}{c}{\textbf{18.26}}          & \multicolumn{1}{c}{\textbf{0.4736}}          & \multicolumn{1}{c}{\textbf{0.4053}}          & \multicolumn{1}{c}{\textbf{0.3705}}          \\ \hline           
\end{tabular}

\label{tab:ablation}
\end{table*}

\subsection{Visual Results}
\subsubsection{Comparison of Optical Images}
The visual results of the various methods are shown in Figure  \ref{fig:visual_results}. The visual results of Pix2pix are blurred, resulting from the lack of explicit use of structure and the lack of constraints on the reconstruction of pixels.
CycleGAN can fully inherit the structural information in SAR images to generate images with clear structures, but these structures are easy to inherit the geometric deformation in SAR images as indicated by the red boxes. At the same time, translation errors are easy to occur in the generated image, such as restoring the building to the texture of the ground in the first image and restoring the river to the texture of the building in the last image.
Although S-CycleGAN can mitigate translation errors to some extent, it has similar phenomena in geometric deformation. It can be seen from Figure \ref{fig:visual_results} that the structural information of FGGAN and EPCGAN networks, compared with S-CycleGAN and CycleGAN, is enhanced. However, the details of the image are still not well recovered. In the results of FGGAN and EPCGAN in the first row, part of the road is lost.
For SAR images with speckle noise, compared with other methods, the proposed S2O-TDN can produce optical images with more detailed structure accuracy and better visual quality.

\begin{figure*}[t!]
	\centering
	\includegraphics[width=1\linewidth]{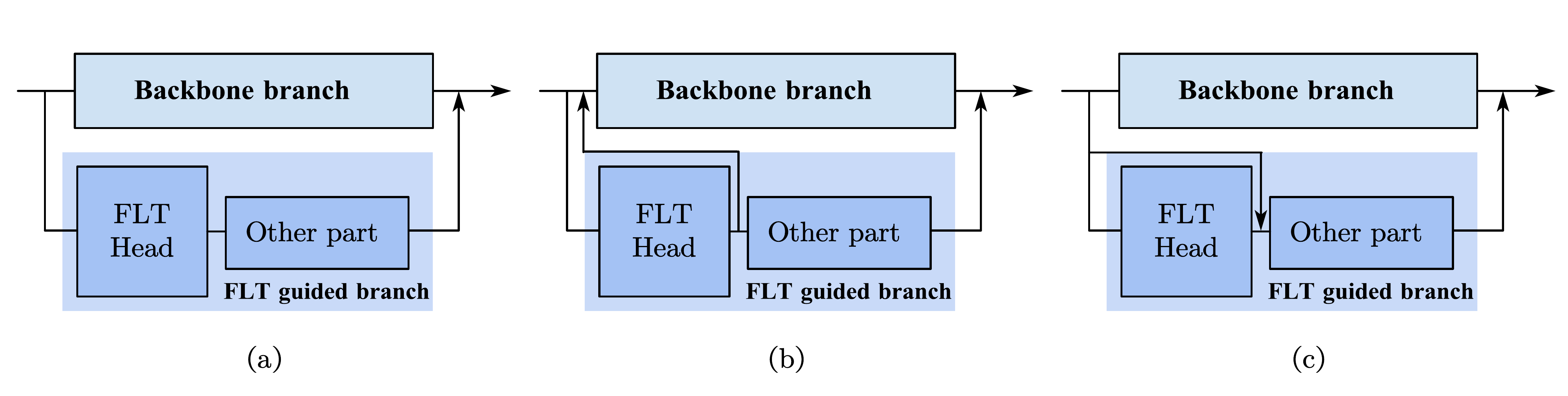}
	\caption{Variant designs of the FLT-guided branch.}
	\label{fig:connection}
\end{figure*}

\subsubsection{Comparison of Structural Information}
The gradient of the image refers to the rate of change of image gray values, which can effectively reflect the texture and structure of an image. We separately select the gradient information of optical images generated from different methods for comparison and seek to demonstrate that texture information and structure information can be retained more effectively in our method. In Figure \ref{fig:visual_structure}, we can find that the texture information is clearly contaminated by the speckle noises in the SAR images, which makes it difficult for the network to extract features to reconstruct the optical images.
As a result, the optical images produced by the currently proposed methods often have problems with geometric distortions and missing textures. Different from them, the proposed S2O-TDN produces better quality results and is closer to ground truth benefiting from the proposed TFD residual blocks and FLT-guided branch.

\subsection{Ablation Study}
\subsubsection{Impact of the Key Components of S2O-TDN}
To investigate the validity of each component in the proposed S2O-TDN, we conduct ablation experiments corresponding to each of its components, as shown in Table \ref{tab:ablation}. 
It is worth stating that the base model does not use the TFD residual block, but the standard residual structure, and the parallel FLT-guided branches,   which do not have any feature interference or auxiliary TD losses.
We try different permutations by addition of the proposed components on the base model separately. It can be seen that both blocks are useful and contribute to achieving better behavior of the model.

Moreover, they are complementary to each other and are used together to produce the best results in the proposed S2O-TDN. In addition, we try different improved residual blocks to demonstrate the superiority of the TFD residual structure. We try alternative designs of TFD residual blocks by using Runge-Kutta2 (RK2)~\cite{he2019ode}, the PolyInception2 (poly2)~\cite{2017PolyNet} and they show inferior performance to our designs on the S2O image translation task. We argue that poly2 is the multiplexing and range expansion of the residual block, which faces difficulties in effectively extracting information disturbed by the noise without specific reorganization of features. In addition, the RK2 block is designed according to the Runge-Kutta solver of ODE, but it has limited representation capability by extracting features only from adjacent layers, differently from the TFD block, it takes features from multi-layers.

\begin{table*}
\normalsize
\centering
\setlength{\tabcolsep}{5mm}
\caption{Results of S2O-TDN with different designs of FLT-guided branch. 
}
\begin{tabular}{ccccccc}
\hline
\multicolumn{1}{c}{\multirow{2}{*}{Method}} & \multicolumn{3}{c}{PSNR}                                                                                       & \multicolumn{3}{c}{SSIM}                                                                                          \\ \cline{2-7} 
\multicolumn{1}{c}{}                        & \multicolumn{1}{c}{Test1}        & \multicolumn{1}{c}{Test2}        & \multicolumn{1}{c}{Test3}        & \multicolumn{1}{c}{Test1}         & \multicolumn{1}{c}{Test2}         & \multicolumn{1}{c}{Test3}         \\ \hline

\multicolumn{1}{c}{S2O-TDN}           & \multicolumn{1}{c}{19.16} & \multicolumn{1}{c}{\textbf{17.94}} & \multicolumn{1}{c}{\textbf{18.26}} & \multicolumn{1}{c}{0.4736} & \multicolumn{1}{c}{\textbf{0.4053}} & \multicolumn{1}{c}{\textbf{0.3705}} \\ \hline

\multicolumn{1}{c}{S2O-TDN(b)}           & \multicolumn{1}{c}{\textbf{19.21}} & \multicolumn{1}{c}{17.74} & \multicolumn{1}{c}{18.18} & \multicolumn{1}{c}{\textbf{0.4827}} & \multicolumn{1}{c}{0.3969} & \multicolumn{1}{c}{0.3659} \\ \hline

\multicolumn{1}{c}{S2O-TDN(c)}           & \multicolumn{1}{c}{18.96} & \multicolumn{1}{c}{17.80} & \multicolumn{1}{c}{17.88} & \multicolumn{1}{c}{0.4505} & \multicolumn{1}{c}{0.3944} & \multicolumn{1}{c}{0.3613} \\ \hline
\end{tabular}

\label{tab:connection}
\end{table*}

\begin{table*}
\normalsize
\centering
\setlength{\tabcolsep}{5mm}
\caption{Model Complexity Analysis of some representative S2O methods and the proposed S2O-TDN.}
\begin{tabular}{cccccc}
\hline
\multicolumn{1}{c}{Method}  & \multicolumn{1}{c}{Pix2pix}        & \multicolumn{1}{c}{CycleGAN}        & \multicolumn{1}{c}{FG-GAN}        & \multicolumn{1}{c}{EPCGAN}         & \multicolumn{1}{c}{\textbf{S2O-TDN}}                \\ \hline

\multicolumn{1}{c}{Paramrers (M)}           & \multicolumn{1}{c}{54.40} & \multicolumn{1}{c}{11.37} & \multicolumn{1}{c}{45.58} & \multicolumn{1}{c}{2.515} & \multicolumn{1}{c}{\textbf{2.063}}  \\ \hline

\multicolumn{1}{c}{FLOPs (G)}           & \multicolumn{1}{c}{17,84} & \multicolumn{1}{c}{56.01} & \multicolumn{1}{c}{62.04} & \multicolumn{1}{c}{64.53} & \multicolumn{1}{c}{\textbf{62.12}}  \\ \hline


\end{tabular}
\label{tab:complexity}
\end{table*}

\subsubsection{Variant Designs of the FLT-guided Branch}
According to Eq.~\eqref{14}, we offer three variants of the FLT-guided branch by using different connections after the FLT head. Figure \ref{fig:connection} shows three different designs. Specifically, Figure \ref{fig:connection}(a) is the default design we proposed in the paper. In the design of Figure \ref{fig:connection}(b), we also add the residual generated by the FLT head back to the input image of the backbone network, \textit{i.e.}, implement the formula derived from FLT at the beginning of the backbone network and keep the role of FLT-guided branch to implement the FLT for the whole network. In Figure \ref{fig:connection}(c), we add the residual generated by the FLT head back to the input image before it is fed into the fusion blocks in the FLT-guided branch, \textit{i.e.}, implement the FLT at the beginning of the FLT-guided branch and the reconstructed information will be integrated with the features from the TFD residual blocks.

The results of the proposed S2O-TDN with the above three designs are summarized in Table~\ref{tab:connection}. As can be seen, the default design and design (b) perform comparably. We hypothesize that the similar performance is since the FLT is both implemented for the whole network in the default design and design (b), \textit{i.e.}, achieving the constraint of feature diffusion and structure preservation for the whole network by the FLT-guided branch. Besides, implementing the FLT at the beginning of the FLT-guided branch will affect the role of the FLT-guided branch, resulting in performance degradation. The experiments imply that leveraging FLT to constrain the entire network rather than individual feature maps by the FLT-guided branch is a better choice.

\subsection{Model Complexity Analysis}
In addition, we analyze the model complexity including the model size, i.e., the number of parameters, and the computation cost, i.e., floating-point operations (FLOPs). The model complexity analysis of the proposed S2O-TDN and some representative S2O methods are summarized in Table ~\ref{tab:complexity}. It should be noted that we count the parameters and FLOPS of individual generators. The smaller model size and better performance of the proposed S2O-TDN compared with other methods indicate that our model has a more effective fitting capability. While obtaining the best performance with respect to high target indicators as well as superior visual quality, the FLOPs of the proposed S2O-TDN are close to those of most existing methods. The reason behinds it is that the FLT-guided branch of the proposed S2O-TDN inevitably introduces additional computation cost.

\section{Conclusion}
In this paper, we introduce the thermodynamic theory to the design of networks for solving the challenging SAR-to-optical image translation task. The proposed S2O-TDN model contains an FLT-guided branch inspired by the first law of thermodynamics and several TFD residual blocks inspired by the third-order finite difference equation, which is used for regularizing the feature diffusion and preserving image structures during S2O image translation, as well as multi-stage feature aggregation processing in translation module. The proposed S2O-TDN can effectively extract the features of SAR images and generate optical images with a clearer and finer structure. Our experiments on the widely used sen1-2 dataset demonstrate that the proposed S2O-TDN has an advantage over most advanced methods in regard to objective indications and visual quality.

\bibliographystyle{IEEEtran}
\bibliography{main}

\end{document}